\documentclass[10pt,twocolumn,letterpaper]{article}

\usepackage[pagenumbers]{cvpr} % To force page numbers, e.g. for an arXiv version

\usepackage{graphicx}
\usepackage{amsmath}
\usepackage{amssymb}
\usepackage{booktabs}

\usepackage[pagebackref,breaklinks,colorlinks]{hyperref}

\usepackage[capitalize]{cleveref}
\crefname{section}{Sec.}{Secs.}
\Crefname{section}{Section}{Sections}
\Crefname{table}{Table}{Tables}
\crefname{table}{Tab.}{Tabs.}

\def\cvprPaperID{91} % *** Enter the CVPR Paper ID here
\def\confName{CVPR}
\def\confYear{2022}

\begin{document}

\title{Rendering Nighttime Image Via Cascaded Color and Brightness Compensation}

\author{Zhihao Li, Si Yi, and Zhan Ma\thanks{Z. Li (lizhihao6@outlook.com) and Z. Ma (mazhan@nju.edu.cn) are with Nanjing University, Jiangsu 210093, China; And S. Yi (1811326@mail.nankai.edu.cn) is with Nankai University, Tianjin, China.} 
%$^\ddag$Nanjing University, $^\dag$Nankai University
}
\maketitle

%%%%%%%%% ABSTRACT
\begin{abstract}
    Image signal processing (ISP) is crucial for camera imaging, and neural networks (NN) solutions are extensively deployed for daytime scenes. The lack of sufficient nighttime image dataset and insights on nighttime illumination characteristics poses a great challenge for high-quality rendering using existing NN ISPs. To tackle it, we first built a high-resolution nighttime RAW-RGB (NR2R) dataset with white balance and tone mapping annotated by expert professionals. Meanwhile, to best capture the characteristics of nighttime illumination light sources, we develop the {\it CBUnet}, a two-stage NN ISP to cascade the compensation of color and brightness attributes. Experiments show that our method has better visual quality compared to traditional ISP pipeline, and  is ranked at the second place in the NTIRE 2022 Night Photography Rendering Challenge~\cite{ershov2022ntire} for two tracks by respective People's and Professional Photographer's choices. The code and relevant materials are avaiable on our website\footnote{https://njuvision.github.io/CBUnet}.
\end{abstract}

%%%%%%%%% BODY TEXT
\section{Introduction}
\label{sec:intro}
% isp 

% main figure
As for camera imaging, photons first converge on the sensor chip to generate the raw electrical image (RAW) reflecting acquired scene; And then the image signal processor (ISP) is usually devised to transform camera RAW image to corresponding RGB representation pleasantly perceivable to human visual system (HVS). Therefore, the efficiency of ISP is of great importance for the generation of high-quality RGB images used in vast applications.
% The main function of Image Signal Processing (ISP) is to post-process the signal output by the front-end image sensor. 
% Depends on the ISP to better restore scene details under different light source. 

Traditional ISP systems usually include demosaicing, automatic white balance (AWB), color space conversion, tone mapping, denoising, compression, etc. Among them, most steps aim at alleviating or eliminating the inherent defects incurred by the image sensor and environment to render high-quality images in terms of the color, brightness, texture sharpness, dynamic range, etc. for visually pleasant presentation to HVS perception. Oftentimes, modular subsystems  in ISP exemplified previously are developed separately, usually taking into account the characteristics of underlying sensor and optics  for systematic optimization.

\begin{figure}[t]
\centering
\begin{minipage}[b]{0.45\textwidth}
\centering
\includegraphics[width=0.9\textwidth]{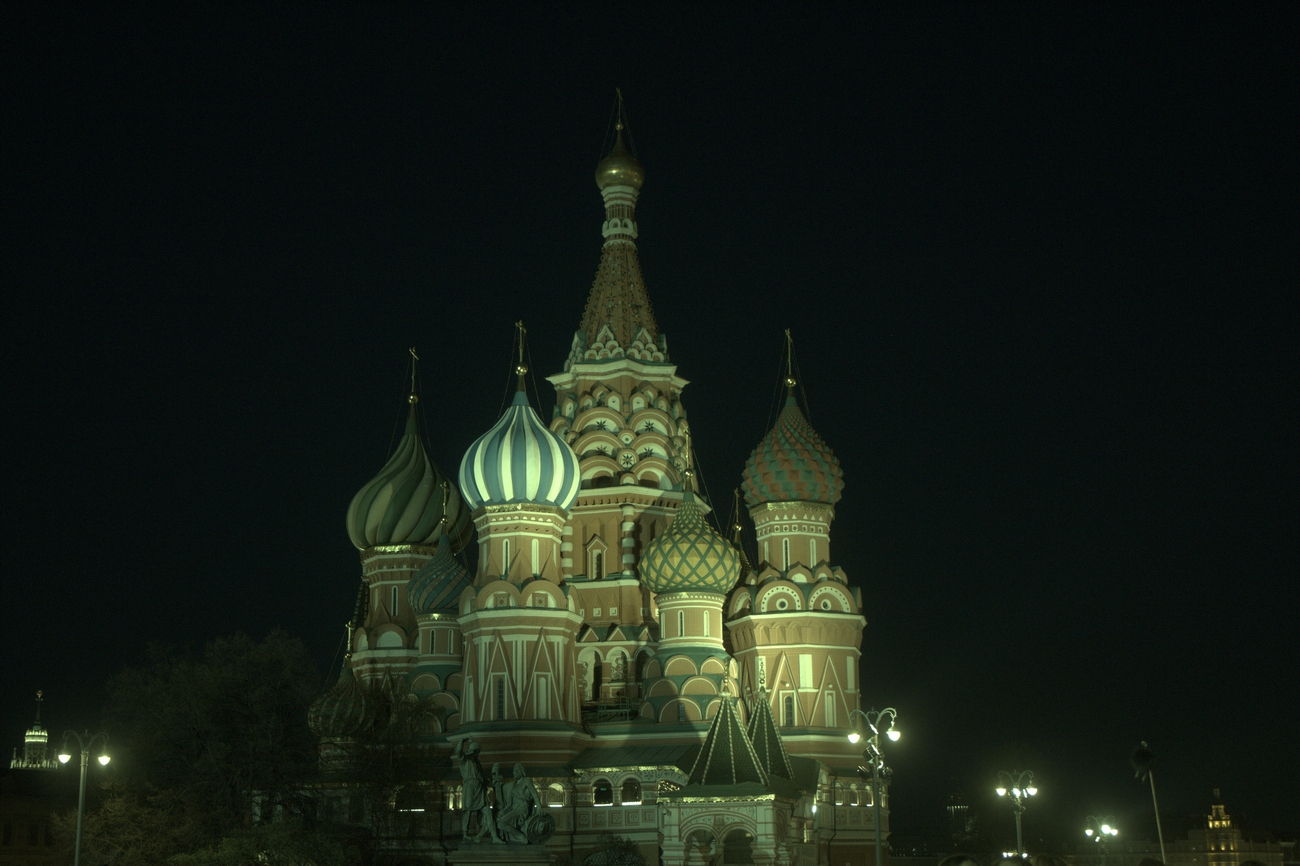}
% \caption{Happy Smiley}
\label{fig:minipage1}
\end{minipage}

\quad

\begin{minipage}[b]{0.45\textwidth}
\centering
\includegraphics[width=0.9\textwidth]{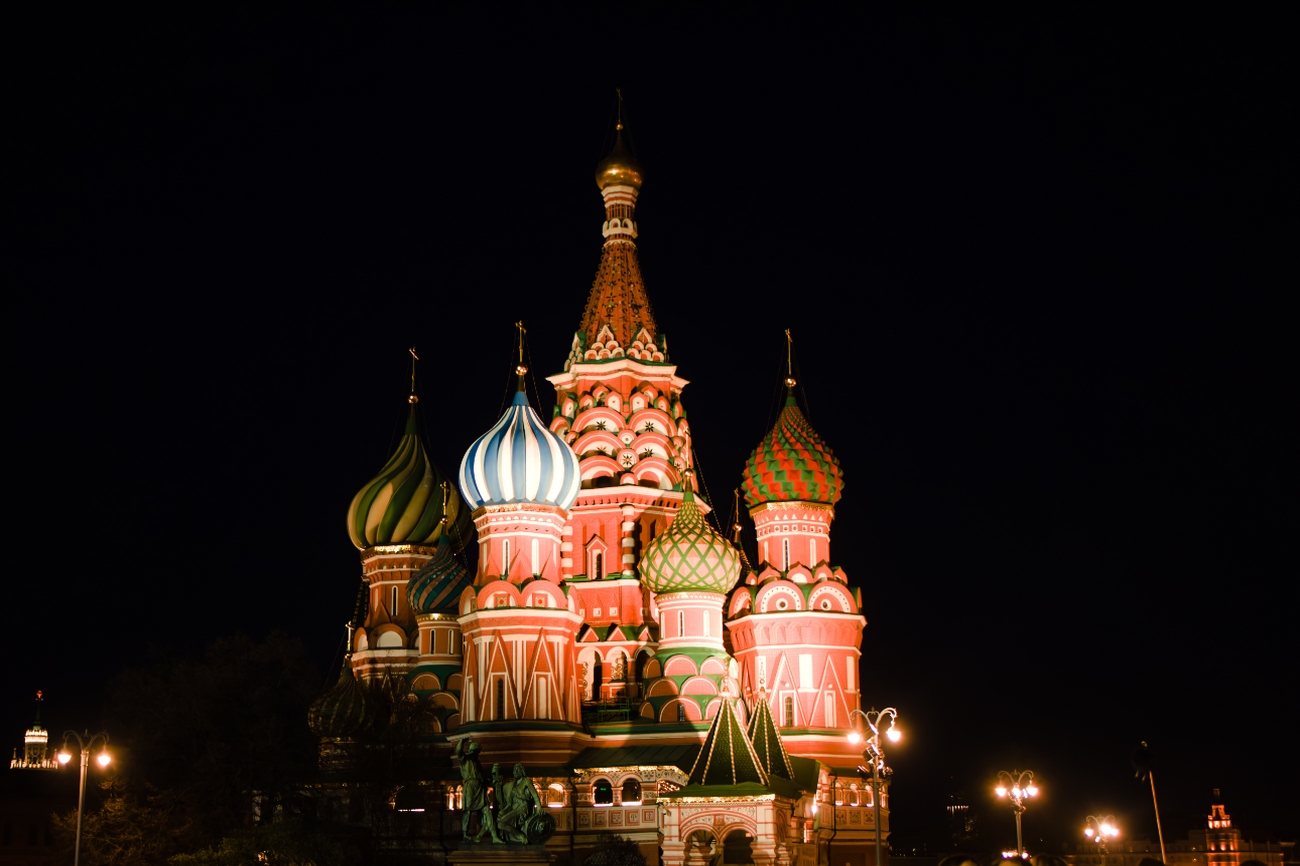}
% \caption{Sad Smiley}
\label{fig:minipage2}
\end{minipage}

\caption{Nighttime RAW image (visualized) and the corresponding RGB image processed rened by the proposed CBUnet.}
\label{main_figure}
\end{figure}

% night image problem (daylight)
In practice,  the HVS can efficiently render dynamic scenes because of the  priors in brain memory~\cite{maloney1999physics,yoshida2005perceptual} (e.g., background illumination, environment understanding), which however is difficult for traditional ISPs to capture due to the lack of understanding of scene. Therefore, some recent studies~\cite{bianco2017single, bianco2019quasi, 9371686} have tried to use neural networks (NN) to replace some modules in ISP to dynamically adjust the output RGB image with scene priors accordingly. For example, Bianco~\etal \cite{bianco2017single} presented a three-stage CNN model method to do illuminant estimation of RAW images. 
% Bianco~\cite{bianco2019quasi} propose a method for computational color constancy in which a deep convolutional neural network is trained to detect achromatic pixels in color images after they have been converted to grayscale.
Panetta~\etal\cite{9371686} proposed a deep learning-based tone mapping operator (TMO-Net), which offered a generalized,  efficient and parameter-free way across a wider spectrum of HDR (high dynamic range) content. Moreover, there had been other works~\cite{ignatov2020replacing, huang2018range, hui2018perception} that attempted to replace the entire ISP system with an end-to-end trainable neural network.

Training a robust NN ISP, e.g., either modular function or end-to-end system, required  a great  amount of paired data, and then significant manpower and resources were devoted to build proper datasets. However,  most exiting datasets were collected at daytime, making it not be applicable for applications in the night because of the great illumination variations between daytime and nighttime. 
% neural network based ISP only trained on daytime images. 
%The difference between daytime and nighttime scenes makes these methods are not fully applicable to the rendering of night images. 
Taking the AWB as an example, daytime light sources are primarily strong sunlight, while nighttime light sources are  more complex, including a variety of artificial light sources that have never appeared in existing daytime datasets. 

To address the lack of paired RAW-RGB data captured at night time, we labeled a high-resolution nighttime RAW to RGB (NR2R) dataset. Specifically, we selected and denoised 150 RAW images with a resolution of 3464$\times$5202 from the training and validation set provided by the night image rendering challenge~\cite{ershov2022ntire}. Then, we first converted the RAW sample to its corresponding RGB format by a simple ISP for extracting the white patch manually. The white patch of each converted RGB image was used to estimate the ground truth illumination for subsequent AWB. Later then, the denoised RAW image was fed through a serial  operations including  bilinear demosaicing, auto white balance with label white balance, color space correction with camera inner color correction matrix (CCM) and neural network based local tone mapping to get a 16-bit intermediate RGB image. Finally we import this 16-bit intermediate RGB image into the Lightroom and manually perform adjustments of global exposure, shadows, highlights and contrast  accordingly to best reflect the visual preference for deriving the final  high-quality 8-bit RGB image.

\begin{table}[t]
    \centering
    \begin{tabular}{l|c|c|c|c}
        \hline
        \hline
        &&& \multicolumn{2}{|c}{Rank} \\
        Team No. & \#Votes & Mean Score  & People & Pro.\\
        \hline
        Team1 & 2603 & 0.8009 & 1 & 1 \\
        \textbf{Ours} & \textbf{2047} & \textbf{0.6298} & \textbf{2} & \textbf{2} \\
        Team3 & 1979 & 0.6089 & 3 & 3 \\
        Team4 & 1964 & 0.6045 & 4 & 4 \\
        Team5 & 1935 & 0.5955 & 5 & 6 \\
        Team6 & 1866 & 0.5742 & 6 & 7 \\
        Team7 & 1559 & 0.4798 & 7 & 8 \\
        Team8 & 1505 & 0.4631 & 8 & 9 \\
        Team9 & 1433 & 0.4411 & 9 & 10 \\
        Team10 & 1288 & 0.3965 & 10 & 5 \\
        \hline
        \hline
    \end{tabular}
    \caption{Result Illustration of Number of Votes (\#Votes), Mean Score, Rank of respective People and Professional Photographer's choice for NTIRE 2022 Night Photography Rendering Challenge.}
    \label{table:challenge_results}    

\end{table}

Unlike daytime images mostly with high illumination, nighttime images were typically acquired under the illumination conditions with complex light sources. We proposed a two-stage  CBUnet to cascade the processing of the color and brightness compensation where  we used a Unet~\cite{ronneberger2015u} with channel-attention for color correction at the first stage, and  in the second stage we applied a histogram-aware Unet for tone mapping by leveraging statistical brightness information from nighttime images. 

Our main contributions can be summarized as follows:
\begin{itemize}
    \item A high-resolution image dataset with nighttime RAW-RGB (NR2R) pairs  that are rendered and annotated by experts with white balance and tone mapping is provided.
    \item A novel two-stage {\it CBUnet} to compensate color and brightness attributes consecutively is developed to render acquired RAW images.
    \item As shown in Table~\ref{table:challenge_results}, our CBUnet achieved the second best performance in both people's and photographer choice of IEEE CVPR NTIRE 2022 Night Photography Rendering Challenge. 
\end{itemize}

\section{Related Work} \label{sec:related_work}
Over past decades, modular subsystems of a camera ISP system had been extensively examined 
like the demosaicing~\cite{li2008image,dubois2006filter,hirakawa2005adaptive}, denoising~\cite{buades2005non,condat2010simple,foi2008practical}, white balancing~\cite{gijsenij2011improving,buchsbaum1980spatial,van2007edge}, tone mapping~\cite{van2007edge,tumblin1993tone,kalantari2017deep}, etc. Most ISP subsystems  had been  successfully emulated and enhanced using deep learning techniques by carefully modeling associated function as the image-to-image translation problems.
% , and their common property is that their goal is to remove noise and artifacts from images.
% The related questions are as follows:

\textbf{Demosaicing} refers to restoring the color information of the remaining two channels of each pixel through interpolation. 
Park~\etal applied residual learning and densely connected convolutional neural network to do color filter array demosaicking, where the proposed model did not require any initial interpolation step for mosaicked input images~\cite{8825809}. 
% Francesco~\etal introduce a novel data-driven model for demosaicing that takes into account the different requirements for reconstruction of the image Luma and Chrominance channels~\cite{deGioia2021DataDrivenCM}.

\textbf{Denoising} aims to remove the noise and recover the latent observation from the given noisy image. Zhang~\etal proposed a depth image denoising and enhancement framework using a lightweight convolutional network~\cite{7472127} where a three-layer network model was applied for high dimension projection, missing data completion and image reconstruction. Zhou~\etal proposed  a novel Bi-channel Convolutional Neural Network (Bi-channel CNN)~\cite{zhou2015learning} for the same purpose.
% It extracts robust representations from raw input by using deep convolutional network, then adaptively integrates two channels of information (the raw input image and face representations) to predict the high-resolution image.

\textbf{White balance} follows the color constancy  of  HVS to eliminate the influence of the color attributes of the light source on scenes acquired by underlying image sensor. Bianco \etal used a CNN model to accurately predict the scene illumination~\cite{bianco2017single} where unlike  handcrafted features explored previously, this CNN model accepted spatial image patches for illumination estimation directly. The CCC~\cite{barron2015convolutional}, a.k.a., convolutional color constancy, and its follow-up refinement FFCC~\cite{barron2017fast}, a.k.a., Fast fourier color constancy, proposed by Barron \etal, reformulated  the color constancy problem as a 2D spatial localization task in log chromaticity space, and thereby applied object detection and structured prediction techniques to solve it.
% FFCC enables better training techniques, an effective temporal smoothing technique, and richer methods for error analysis.

\textbf{Tone mapping} operator converts High Dynamic Range (HDR) image to its Low Dynamic Range (LDR) representation for the rendering on normal LDR displays. To train HDR-LDR mapping, a popular approach  was to use a tone-mapping operator (TMO) to generate a set of image labels and narrow it down using an image quality index to obtain final training examples~\cite{patel2017generative,rana2019deep}. By incorporating the manual supervision, Zhang~\etal proposed a tone mapping network (TMNet) in Hue-Saturation-Value (HSV) color space to obtain better luminance and color mapping results~\cite{zhang2019deep}.  Montulet \etal introduced a new end-to-end tone mapping approach based on Deep Convolutional Adversarial Networks (DCGANs) along with a data augmentation technique, which reportedly showed the state-of-the-art results on benchmark datasets~\cite{montulet2019deep}.

As a matter of fact, most works mentioned above mainly paid their attention on  daytime image optimization. In contrast, this work deals with the nighttime image rendering by building a nighttime RAW-RGB dataset and developing a two-stage  CBUnet to reconstruct ``visually more pleasing'' images from  RAW scenes acquired during the night.

\section{Method}

\subsection{NR2R Dataset}

\begin{figure*}[ht]
\centering
\includegraphics[width=0.8\textwidth]{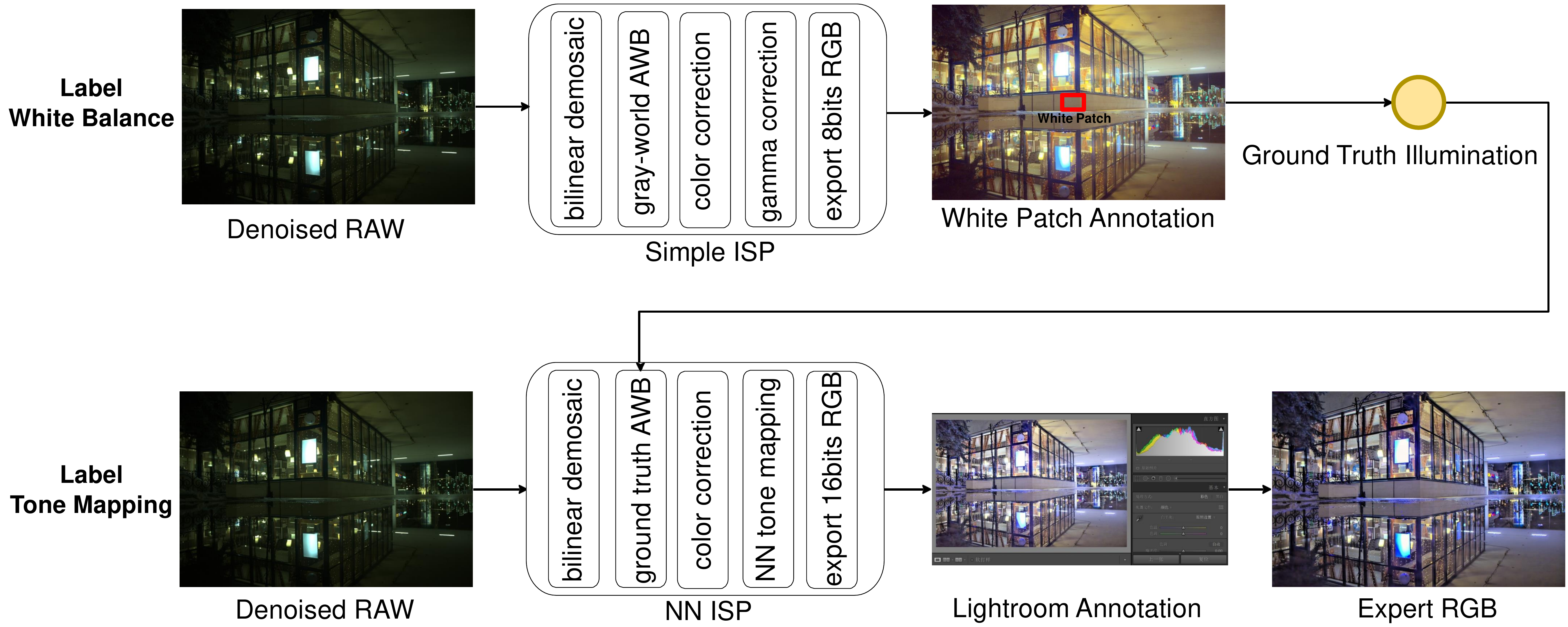}
\caption{Labeling pipeline. We first perform the white balance annotation to derive the ground truth illumination;  then input denoised RAW with estimated ground truth illumination into the NN ISP module (including bilinear demosaic, AWB, color correction, tone mapping) to obtain the brightness-adjusted RGB image, and finally use Lightroom to adjust global exposure, brightness, shadows and contrasts.}
\label{label}
\end{figure*}

{\bf RAW Samples.} To form the collection of nighttime RAW samples, we first selected a total of 150 images with the spatial resolution at 3464$\times$5202 from the training and validation sets provided by the night image challenge~\cite{ershov2022ntire}. 
%\item
%\end{item}
And then these RAW images are pre-processed to best produce noise-free samples using a notable CNN based denoiser~\cite{abdelhamed2020ntire}. This is because nighttime imaging experiences a very challenging situation with heavy noises incurred by high ISO setting under poor illumination condition (e.g., underexposure).

%the captured nighttime image contain a lot of noise due to the high ISO within underexposure condition. 
{\bf Paired RGB Derivation.} We applied a two-stage process to derive the corresponding RGB image of each RAW input.

As shown in Fig.~\ref{label}, we first used a simple ISP that was comprised of linear demosaicing, gray-world white balance, color correction, and gamma correction to convert each denoised RAW input to its RGB format for groundtruth illumination estimation. To this aim,  we mark the ``White Patch'' from each converted RGB, where the patch is presented in neutral gray, and its RGB channels are approximately the same. Since the gray surface presumably reflects all incoming light radiation, it can be used to represent the ground truth illumination of the RAW image accordingly. 

We  then perform the 2-stage labeling using the illumination from the 1-stage.  Specifically, first we get the correct color image by a serial operations including linear demosaicing, white balance using the label white balance and color correction with the camera inner color correction matrix (CCM). The brightness adjustment consists of local and global tone mapping jointly. Since local tone mapping requires fine-grained adjustment of each small patch in the scene, it is difficult to annotate it manually. Therefore, we use a pre-trained local tone mapping model in~\cite{vinker2021unpaired} to fulfill the task. Since the pre-trained tone mapping network was trained using daytime image, it is good for local adjustment, but fails to control the global brightness.  We save the model output 
using a 16-bit intermediate format in PNG, and then  import it into the Lightroom app to adjust the global exposure, brightness, shadows and contrast manually for final high-quality RGB image rendering, with which we emulate the image rendering knowledge from Professional Photographers.

Thereafter, we successfully obtain a high-resolution nighttime RAW-RGB image dataset for the training of CBUnet in next sections.

\subsection{CBUnet Architecture}

\begin{figure*}
\centering
\includegraphics[width=0.8\textwidth]{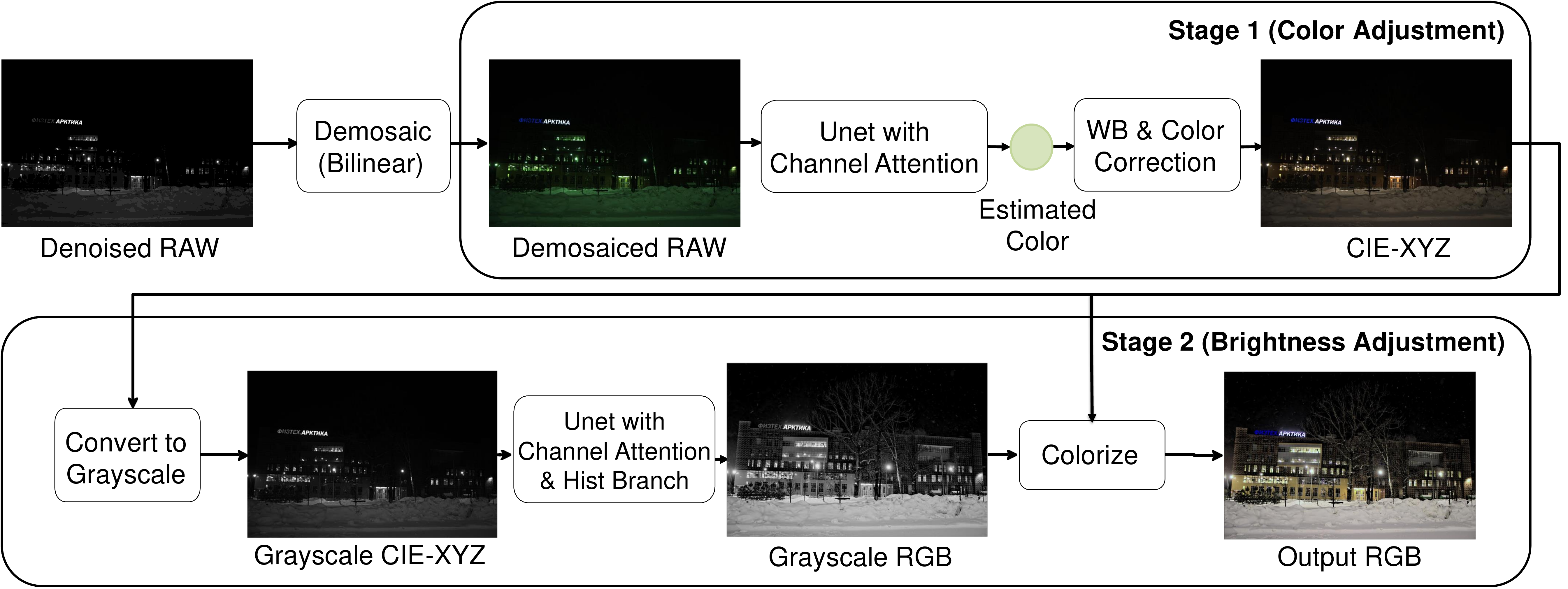}
\caption{The architecture of the CBUnet. It consist of two stages, where the first stage is used to correct the color and the second one adjusts the brightness for a high-quality output.}
\label{two-stage}
\end{figure*}

\begin{figure*}
\centering
\includegraphics[width=0.8\textwidth]{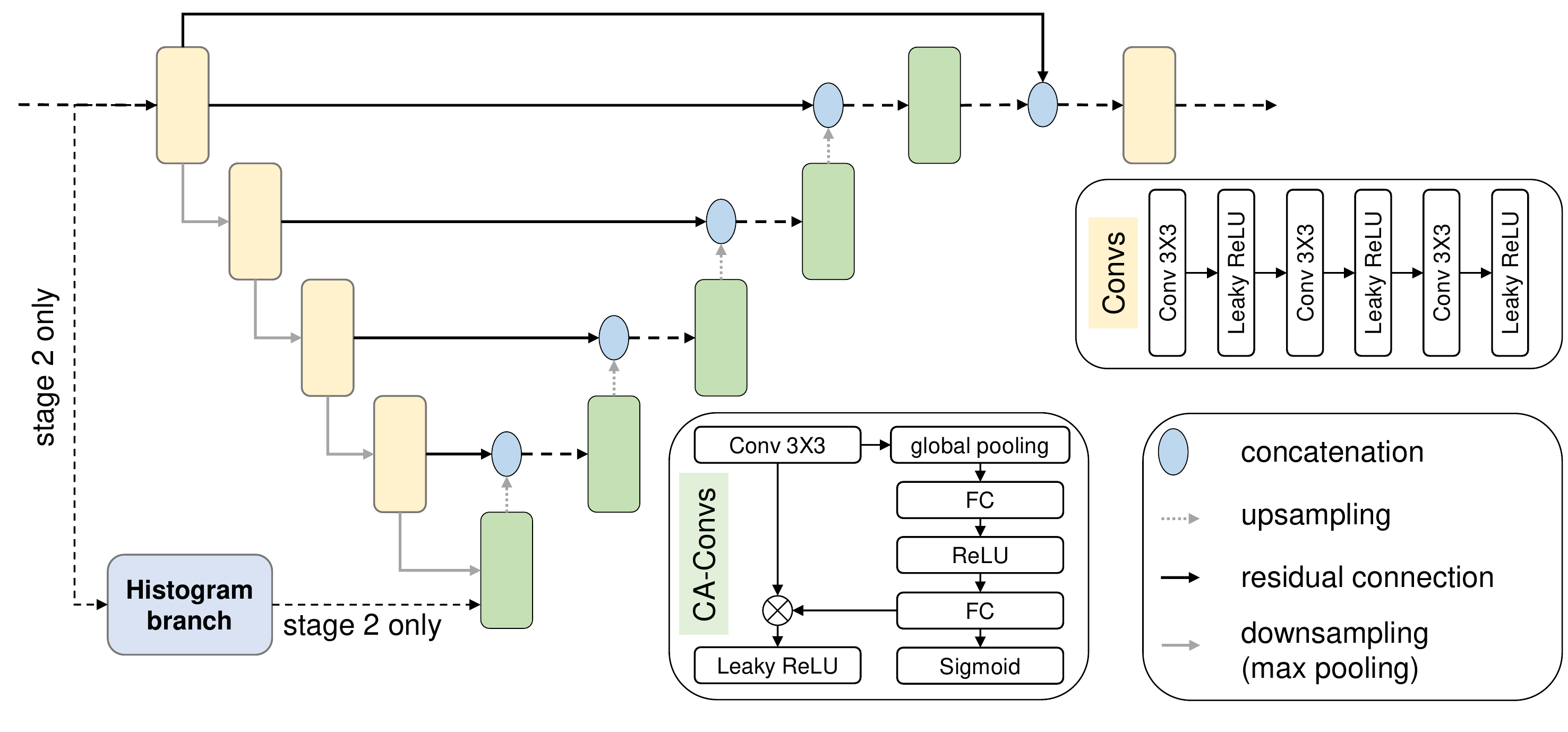}
\caption{The architecture of our modified Unet. The histogram branch is only used in stage 2.}
\label{caunet}
\end{figure*}

Figure~\ref{two-stage} illustrates the architecture of the proposed CBUnet. It consists of two stages, which are cascaded for color correction and brightness adjustment, respectively. The first stage takes an demosaiced noise free RAW image $I_{raw}$ as input. We modified an encoder-decoder based Unet~\cite{barron2015convolutional} as our primary backbone to estimate the illumination color. The network's deep features are usually too generic without special emphasis even though it is clear that some channel features are more significant than others in different scenes. In order to make the Unet pay more attention to these crucial channels, we employ channel attention-convolutions (CA-Convs) blocks inherited from W-Net~\cite{w_net} to specify the various scenes' response. As shown in Figure~\ref{caunet}, CA-Convs block first uses global pooling to extract spatial information from convolutional features, and then transforms them via fully connected layers (FC), ReLU, and sigmoid. At last, it multiplies the convolutional features with sigmoid's output, a.k.a., the weights of the channel attention. All activation functions are implemented with 0.2 negative slope's parametric rectified linear units (PReLU)~\cite{prelu}. We apply global average pooling to get the predict illumination color $\hat{R}$ at the output layer. The input $I_{raw}$ is then corrected based on the $\hat{R}$, and the color space is further converted to CIE-XYZ based on the in-camera CCM. The whole process of stage 1 can be written as:
\begin{equation}
    \hat{I}_{cie\raisebox{0mm}{-}xyz} = CCM \cdot f\left( I_{raw} \right) \cdot I_{raw},
\end{equation}
where $f\left( \cdot \right)$ is the illumination color estimation neural network.

The second stage is mainly used to adjust the brightness, so we do not change the color properties of the image at this stage. We first extract the grayscale component of $\hat{I}_{cie\raisebox{0mm}{-}xyz}$ and then feed it into the brightness prediction network $g\left(\cdot\right)$ to get the target brightness map, and then colorize it according to the color information in $\hat{I}_{cie\raisebox{0mm}{-}xyz}$. The structure of the $g\left(\cdot\right)$ is similar to the $f\left(\cdot\right)$ in stage 1, but an additional histogram extraction branch is added to obtain the global brightness distribution. Histogram extraction branch uses a 256 bits histogram of the $\hat{I}_{cie\raisebox{0mm}{-}xyz}$  as input, passes through two fully connected layers with ReLU activation functions, and finally expands to the same size as Unet's bottom feature and sums directly. Thus, the stage 2 could be formulated as:
\begin{equation}
    \hat{I}_{rgb} = \frac{ g\left( \textbf{G} \left(  \hat{I}_{cie\raisebox{0mm}{-}xyz} \right) \right) }{ \textbf{G} \left(  \hat{I}_{cie\raisebox{0mm}{-}xyz} \right) } \cdot \hat{I}_{cie\raisebox{0mm}{-}xyz},
\end{equation}
where $\textbf{G}\left(\cdot \right)$ is the grayscale function.

\section{Experiment}
\subsection{Loss Function}
\textbf{Angular Loss} 
In stage 1, we use angular loss as an evaluation criterion between prediction illumination color $\hat{R}$\cite{bianco2017single, bianco2019quasi} and ground truth illumination $R$:
\begin{equation}
    L_{angular} = \frac{180}{\pi} arccos(\hat{R}\cdot R).
\end{equation}

\noindent\textbf{Pixel Loss}
In stage 2, we first use L1 Loss to ensure the accuracy of local tone mapping, which is defined as:
\begin{equation}
    L_1 = \Vert \mathbf{G} ( \hat{I}_{rgb} )  - \mathbf{G} ( I_{rgb} ) \Vert.
\end{equation}

\textbf{Histogram Loss}
However, it is difficult to constrain the global brightness distribution by using $L_1$ only, so we use the histogram loss to make the histogram of the generated images conform to the statistical distribution of the nighttime images. The histogram loss could be formulated as:
\begin{equation}
    L_{hist} = \Vert \mathbf{H} ( \hat{I}_{rgb} )  - \mathbf{H} ( I_{rgb} ) \Vert,
\end{equation}
where $\mathbf{H}$ is differentiable histogram function from~\cite{ustinova2016learning}.

Finally, we define our loss function by the sum of the
aforementioned losses as follows:

\begin{equation}
    L_{total} = L_{angular} + L_1 + L_{hist}.
\end{equation}

\subsection{Training Details}
The model was implemented in Pytorch and was trained on a single Nvidia Tesla 3090 GPU with a batch size 16. We devide the NR2R dataset into two parts: 120 images are used for training and the rest 30 images are used for testing. The stage 1 was first pretrained on Cube++ dataset~\cite{9296220} which is captured within the same camera of the NR2R.Then we use Adam optimizer~\cite{kingma2014adam} with 5e-5 learning rate. Then the stage 2 was trained for 300 epochs with the same learning rate while the parameters of stage 1 was frozen. Finally, the stage 1 and stage 2 were joint finetuned for 10 epochs.

\subsection{Results}
\begin{table}[]
\centering
\begin{tabular}{l|c|c|c}
\hline
\textbf{Method} & \textbf{PSNR}  & \textbf{Parameter (M)} & \textbf{FLOPs (G)} \\ \hline
PyNET\cite{pynet}           & 20.67          & 47.55                  & 1370.79            \\ \hline
HERN\cite{hern}            & 19.57          & 56.18                  & 466.74             \\ \hline
% W-Net           & 21.03          & 33.38                  & 626.57             \\ \hline
AWNet\cite{awnet}          & 21.16          & 46.99                  & 1532.46            \\ \hline
\textbf{Ours}   & \textbf{22.29} & \textbf{23.64}         & \textbf{391.58}    \\ \hline
\end{tabular}
\caption{Comparative studies of reconstruction PSNR, parameter size and FLOPs, where FLOPs is calculated when the input size is $1024\times1024$.}
\label{tab:compare_methods}
\end{table}

\begin{figure*}
\centering
\includegraphics[width=0.9\textwidth]{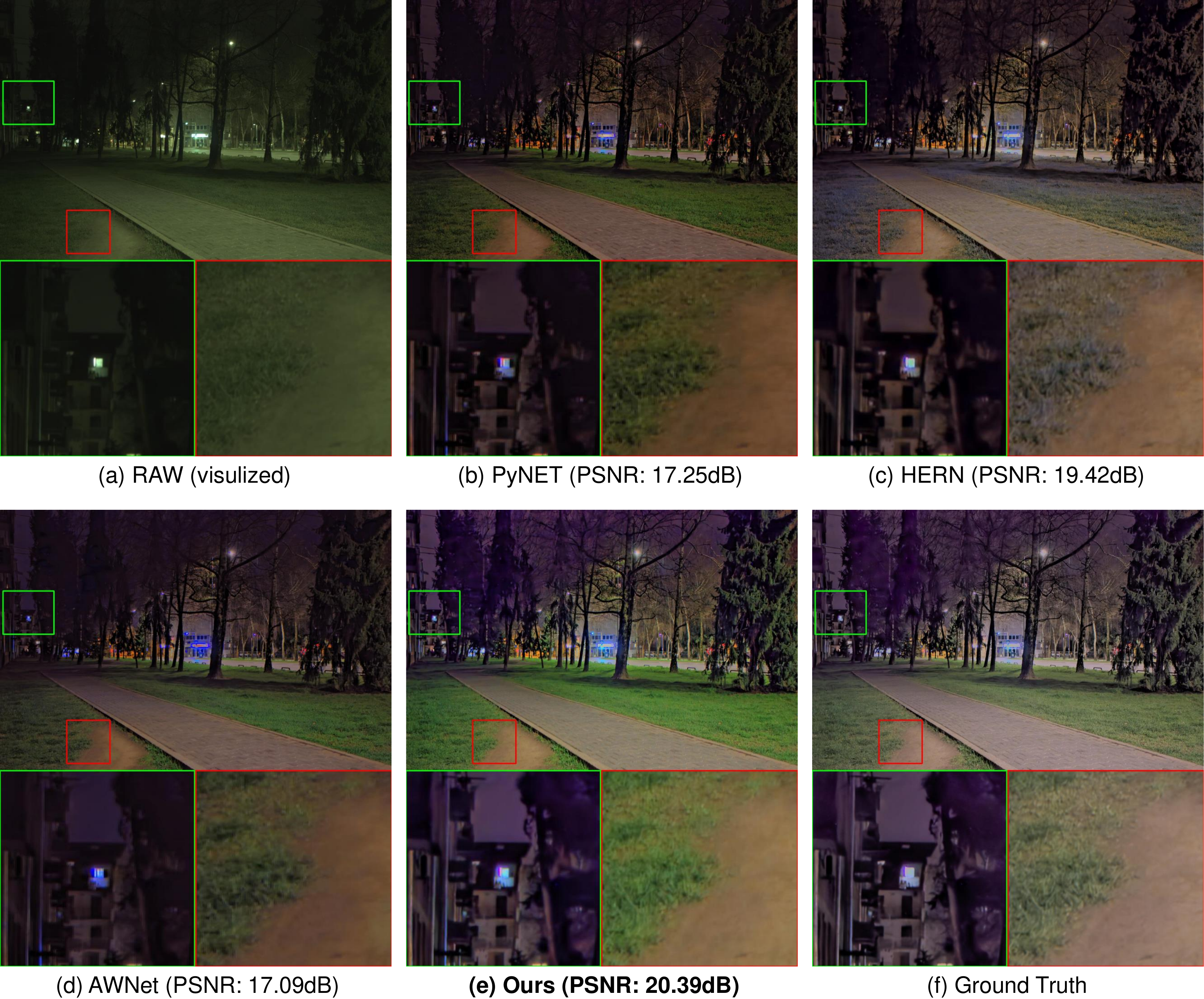}
\caption{Visual comparison of reconstructions from four different methods including PyNET~\cite{pynet}, HERN~\cite{hern}, AWNet~\cite{awnet} and our proposed CBUnet.}
\label{fig:combine_methods}
\end{figure*}
We compare our CBUnet with previous learnt approaches like PyNET~\cite{pynet}, HERN~\cite{hern} and AWNet~\cite{awnet}. As shown in Table~\ref{tab:compare_methods}, we list the reconstruction PSNR (averaged), parameters and floating-point operation (FLOPs). Our CBUnet offers the state-of-the-art efficiency for all metrics. More importantly, our method provides 1.14dB gains over AWNNet~\cite{awnet}, but its FLOPs and parameters only present a very small fraction percentage (e.g., $<$26\%). In the mean time, as shown in Figure~\ref{fig:combine_methods}, our method have demonstrated the superior performance to other solution in synthesing the local details, color saturation and contrast.

\subsection{Ablation Study}
{\bf Network Architecture} First, we demonstrated the effectiveness of our CBUnet. As shown in Table~\ref{tab:ablation_arch}, our most primitive network is Unet. The channel attention brought a gain of 0.68 dB. The two stage design is the most important part of our network performance improvement. The addition of two stage brought 1.48 dB performance gain. Finally, the addition of histogram branch to extract global information results in a gain of 0.28 dB.

\begin{table}[]
\centering
\begin{tabular}{c|c|c|c}
\hline
\textbf{CA} & \textbf{Two Stage} & \textbf{Hist Branch} & \textbf{PSNR}  \\ \hline
            &                    &                           & 19.85 \\ \hline
$\surd$     &                    &                           & 20.53 \\ \hline
$\surd$     & $\surd$            &                           & 22.01 \\ \hline
$\surd$     & $\surd$            & $\surd$                   & \textbf{22.29} \\ \hline
\end{tabular}
\caption{Ablation studies on network architectures, where CA and Hist Branch mean channel attention and histogram branch, respectively.}
\label{tab:ablation_arch}
\end{table}

{\bf Loss Function} Then, we verified the efficiency of our loss function which consists of pixel loss $L_1$, angular loss $L_{angular}$ and histogram loss $L_{hist}$. The experimental results are shown in Table~\ref{tab:ablation_loss}. When $L_{angular}$ is added, 0.16 dB PSNR increase was achieved. Finally, the addition of $L_{hist}$ also improved the results in a gain of 0.15 dB.
\begin{table}[]
\centering
\begin{tabular}{c|c|c|c}
\hline
${L_{1}}$ & ${L_{angular}}$ & ${L_{hist}}$ & \textbf{PSNR}  \\ \hline
$\surd$     &                    &                           & 21.98 \\ \hline
$\surd$     & $\surd$            &                           & 22.14 \\ \hline
$\surd$     & $\surd$            & $\surd$                   & \textbf{22.29} \\ \hline
\end{tabular}
\caption{Ablation studies on loss functions.}
\label{tab:ablation_loss}
\end{table}

\subsection{Generalization to Other Camera Sensors}
\begin{figure*}
\centering
\includegraphics[width=0.9\textwidth]{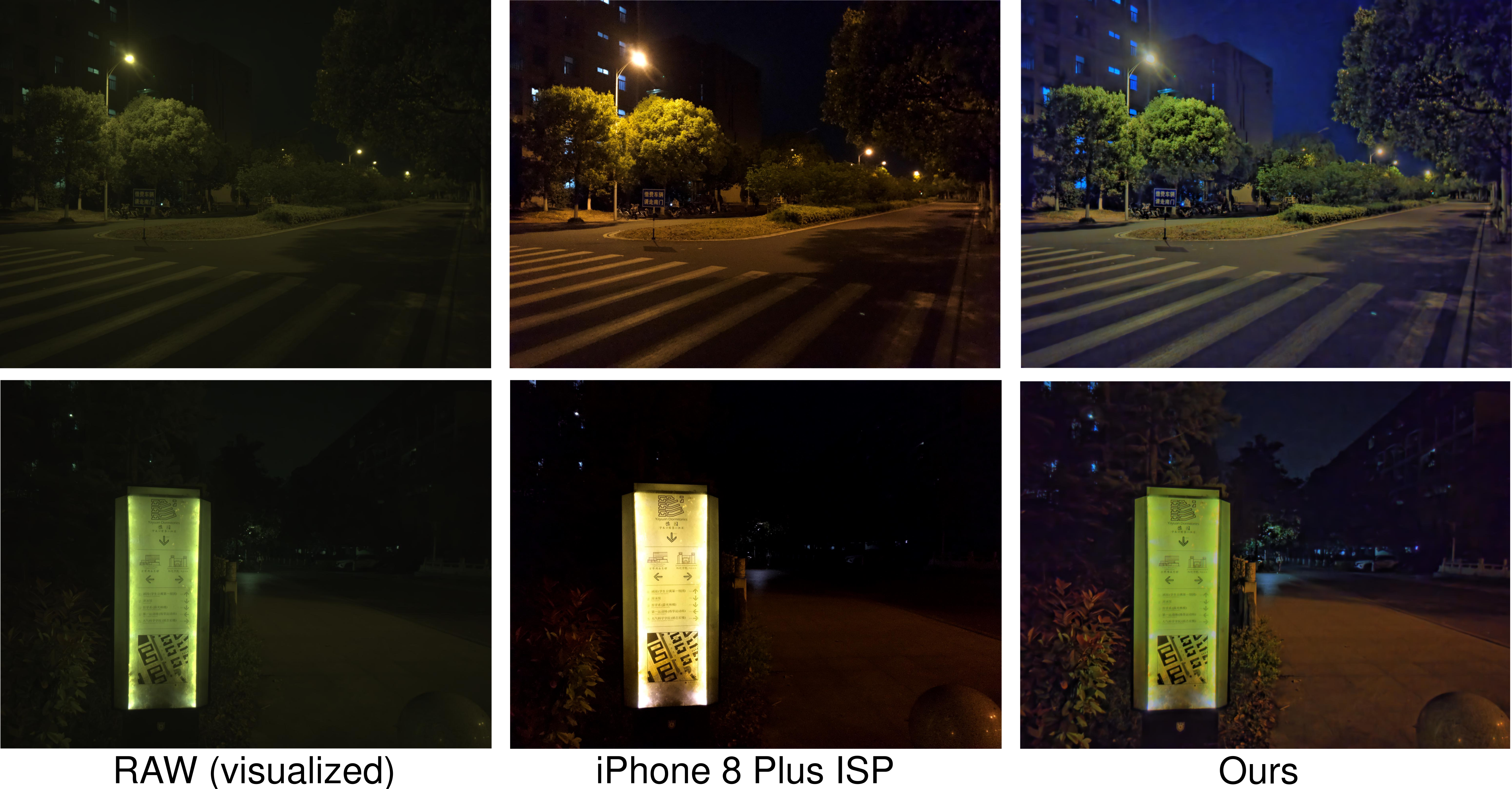}
\caption{The results of the proposed CBUnet on RAW images from the iPhone 8 Plus smartphone. From left to right: the original
visualized RAW image, the same photo obtained with iPhone’s built-in ISP system and our reconstructed RGB image.}
\label{fig:iphone}
\end{figure*}

Although our proposed CBUnet achieved the best performance with minimal computation in the NR2R dataset, but the NR2R dataset was captured by a single DSLR camera. Thus, we test our method on a mobile phone camera to verify the generalization of our method. We took a set of nighttime RAW images with a resolution of $4032\times3024$ using the main camera of iPhone 8 Plus which has a completely different CMOS and optical lens. Figure~\ref{fig:iphone} shows that our CBUnet has better color correction compared to the iPhone's internal ISP, e.g. zebra lines are correctly corrected to white and the sky is blue. Also, our model renders images with better retention of details in the shadow and better suppression of highlights such as haloes.

\subsection{NTIRE 2022 Night Photography Rendering Challenge}
\begin{figure*}
\centering
\includegraphics[width=0.9\textwidth]{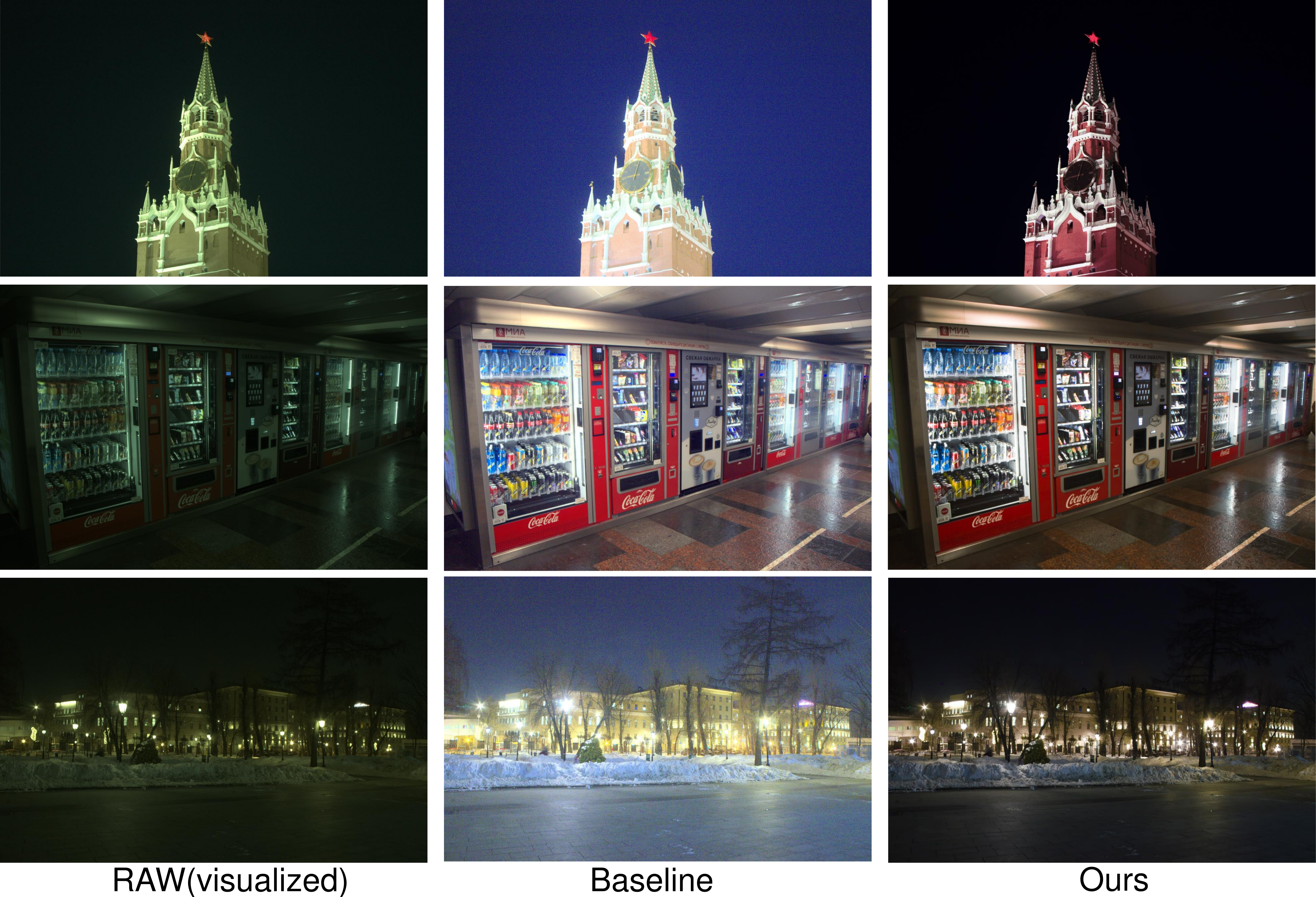}
\caption{Rendering Results of our CBUnet in the NITRE 2022 Photography Rendering Challenge~\cite{ershov2022ntire}.}
\label{results}
\end{figure*}
As for the NTIRE 2022 Night Photography Rendering Challenge~\cite{ershov2022ntire}, We submitted 100 night images rendered with this method.  Rendering results are shown in Figure~\ref{results}.
% The final result is based on the people's opinion and professional photographer's opinion. The leaderboard is formed by including every submission in 3250 comparisons using the Yandex Toloka platform. The votes score mean multiplied by 3250 gives the number of actual votes that a particular submission received.in both leaderboard, we are ranked second place. Qualitative results are shown in Figure ~\ref{results}.
The baseline ISP provided by challenge organizers consists of bilinear demosaicing, bilateral denosing, AWB using gray world assumption, color space conversion with camera inner CCM and flash tone mapping~\cite{banic2018flash}. Compared to the baseline ISP, our renderings have a more accurate white balance and a brightness distribution that is more consistent with nighttime scene characteristics.

\section{Conclusion}
Night photography is challenging due to the lack of sufficient nighttime image dataset and comprehensive understanding of complex light illumination in the night. This work therefore built a NR2R dataset with dedicated expert annotations as the ground-truth for NN ISP training, and developed a { CBUnet} for rendering image in the night.  The proposed { CBUnet} showed high-performance and consistent imaging capacity voted by both experts and amateurs, reporting the second place in the NTIRE 2022 Night Photography Rendering Challenge for both People's and Photographer's choices.

%our method has better visual quality compared to traditional ISP pipeline, and  is ranked at the second place in the NTIRE 2022 Night Photography Rendering Challenge\footnote{https://nightimaging.org/final-leaderboard.html} for two tracks by respective People's and Professional Photographer's choices.

%%%%%%%%% REFERENCES
{\small
\bibliographystyle{ieee_fullname}
\bibliography{egbib}
}

\end{document}